\documentclass[sigconf,nonacm]{acmart}
\usepackage{amsmath}
\usepackage{graphicx}
\usepackage{algpseudocode}
\usepackage{algorithm}
\usepackage[compact]{titlesec}
\usepackage{gensymb}
\usepackage{balance}

\AtBeginDocument{%
  \providecommand\BibTeX{{%
    \normalfont B\kern-0.5em{\scshape i\kern-0.25em b}\kern-0.8em\TeX}}}

\setcopyright{acmcopyright}
\copyrightyear{2022}
\acmYear{2022}
\acmDOI{XX.XXXX/XXXXXXX.XXXXXXX}

\acmConference[SIGKDD '22]{SIGKDD '22: ACM SIGKDD Conference on Knowledge Discovery and Data Mining}{Aug 14--18, 2022}{Washington DC}
\acmBooktitle{SIGKDD '22: ACM SIGKDD Conference on Knowledge Discovery and Data Mining,
  Aug 14--18, 2022, Washington DC}
\acmPrice{XX.XX}
\acmISBN{978-1-4503-XXXX-X/18/06}

\begin{document}

\fancypagestyle{firstpagestyle}
{
   \fancyhf{}
   \setlength{\footskip}{7mm}
   \fancyfoot[L]{\footnotesize 2nd International Workshop on Online and Adaptive Recommender Systems, 28th ACM SIGKDD Conference on Knowledge Discovery and Data Mining, 2022, Washington DC.}
}
\thispagestyle{firstpagestyle}

\title{Dynamic Memory for Interpretable Sequential Optimisation}

\author{Srivas Chennu}
\authornote{Authors contributed equally to the paper}
\email{srivas.chennu@apple.com}
\affiliation{%
  \institution{Apple}
  \country{}
}
\author{Andrew Maher}
\authornotemark[1]
\email{andrew_maher@apple.com}
\affiliation{%
  \institution{Apple}
  \country{}
}
\author{Jamie Martin}
\authornotemark[1]
\email{jamiemartin@apple.com}
\affiliation{%
  \institution{Apple}
  \country{}
}
\author{Subash Prabanantham}
\authornotemark[1]
\email{subash@apple.com}
\affiliation{%
  \institution{Apple}
  \country{}
}

\renewcommand{\shortauthors}{Chennu et al.}

\begin{abstract}

Real-world applications of reinforcement learning for recommendation and experimentation faces a practical challenge: the relative reward of different bandit arms can evolve over the lifetime of the learning agent. To deal with these non-stationary cases, the agent must forget some historical knowledge, as it may no longer be relevant to minimise regret. We present a solution to handling non-stationarity that is suitable for deployment at scale, to provide business operators with automated adaptive optimisation. Our solution aims to provide interpretable learning that can be trusted by humans, whilst responding to non-stationarity to minimise regret. To this end, we develop an adaptive Bayesian learning agent that employs a novel form of dynamic memory. It enables interpretability through statistical hypothesis testing, by targeting a set point of statistical power when comparing rewards and adjusting its memory dynamically to achieve this power. By design, the agent is agnostic to different kinds of non-stationarity. Using numerical simulations, we compare its performance against an existing proposal and show that, under multiple non-stationary scenarios, our agent correctly adapts to real changes in the true rewards. In all bandit solutions, there is an explicit trade-off between learning and achieving maximal performance. Our solution sits on a different point on this trade-off when compared to another similarly robust approach: we prioritise interpretability, which relies on more learning, at the cost of some regret. We describe the architecture of a large-scale deployment of automatic optimisation-as-a-service where our agent achieves interpretability whilst adapting to changing circumstances.

\end{abstract}


\maketitle

\section{Introduction}
Sequential optimisation is the process of progressively learning from data to make better decisions over time. In the machine learning literature, it is commonly framed as the multi-armed bandit problem, where a reinforcement learning agent progressively learns with the aim of maximising overall reward. Such agents have been applied to build experimentation \cite{Dimmery2019ShrinkageExperiments} and recommendation systems \cite{Li2010ARecommendation}. However, in real-world deployments of such learning agents, there is an additional challenge: the relative reward of different bandit arms evolve over the lifetime of the learning agent. How do we ensure that the agent can adapt to a changing environment? To be able to address this difficult challenge of learning under non-stationary conditions, the agent has to not only remember the recent knowledge it has gained, it also has to decide which historical knowledge to forget, as this knowledge might no longer be relevant to minimise regret. This decision to appropriately forget is essential for the agent to be able to adapt to changes in rewards. 

Multiple mechanisms have been proposed for such forgetting, which aim to maximise the agent’s long-term yield in the face of changing circumstances. However, these mechanisms are often not designed with interpretability in mind. Many bandit applications focus solely on the problem of minimising regret, without placing consideration on the concomitant human interpretability of its behaviour. In any real-world system in which humans engage with the output of machine intelligence, such interpretability must be at the forefront. Further, many current proposals for handling non-stationarity typically require an explicit specification of the memory that the agent should retain, for example via a fixed time window \cite{Trovo2020Sliding-WindowSettings} or a prescribed discounting factor \cite{Raj2017TamingApproach}. Although sensible values for these parameters can be identified through hyperparameter optimisation, it is difficult to ensure that a particular value will perform well across the range of non-stationarities that a real-world system might encounter.

To overcome this limitation, we propose a solution to the problem of adaptation to changing rewards that allows memory to grow and shrink dynamically as and when it is needed. We develop an adaptive Bayesian learning agent algorithm that employs a novel form of dynamic memory of historical rewards. A particular feature of our approach is that it actively enables interpretability with statistical hypothesis testing, by targeting a desired set point of statistical power when comparing rewards. We implement robust hypothesis testing using sequential Bayes factors that track statistical confidence. More specifically, whenever the agent receives a new batch of data, it computes statistical confidence about the relative difference between arm rewards, based on the data it has observed thus far. When the desired level of level of statistical confidence is reached – indicating either that arm rewards are significantly different or that they are identical – the agent progressively forgets its knowledge of historical rewards, thereby reducing its memory. If statistical confidence drops below the desired level, the agent begins to re-grow its memory of historical rewards until statistical confidence is restored.

Using numerical simulations, we compare the performance of our solution to the problem of forgetting against the ADWIN algorithm \cite{Bifet2007LearningWindowing}, an existing adaptive memory solution. We show that under a range of tractable examples of non-stationarity, our agent achieves a balance between stability against noise, adaptivity to real changes in the true rewards, and interpretability of this adaptation. A human operator can control our adaptive Bayesian agent by setting a desired level of statistical power, thereby enabling them to interpret the agent’s knowledge and make more effective data-driven business decisions. To show this, we describe an overall system architecture that enables real-world deployment of optimisation at scale. We visualise the performance of a deployed learning agent, and demonstrate its adaptive behaviour.

\section{Background}

Our contribution connects two challenges in sequential optimisation, \emph{interpretability} and \emph{non-stationarity}, introduced in the following sections.

\subsection{Interpretable Sequential Optimisation}
Sequential optimisation refers to the process of progressively learning about the value of a set of alternative actions our learning agent can take. One well-known application of this in A/B testing, where we just want to measure statistically significant differences between a set of variants or treatments like in a randomised controlled trial. In A/B testing, the agent assigns treatment units (e.g., clients of an online service) randomly, seeking only to learn about the alternatives, rather than trying to maximise its reward. A complementary perspective comes from research into multi-armed bandits, where the agent aims to adapt the selection of alternatives over time so as to maximise long-term reward. While both of these approaches have received a lot of attention, they tend to be seen as separate problems.

We adopt an intermediate perspective: can we achieve a balance of both objectives, i.e., maximise yield on the value we care about with reinforcement learning, while also conducting statistical analysis of the alternatives? Obviously, there is a trade-off here: selecting all variants equally would generate a lot of information about each of them, but we would lose out on the yield we would get if we picked the best one all the time. Here, we propose that it is possible to learn only as much as is desired before yielding, finding a desirable sweet spot in the middle of this trade-off. In particular, we will focus on using Bayesian inference to achieve this balance.

Why do we want to achieve this trade-off? We want to facilitate optimisation that has humans in the loop. Human operators are often involved in how optimisation is applied in industry. These experts are involved in business-relevant interpretation, higher-order decision making and meta learning about what delivers value in the long run. For example, human operators might want to use hypothesis testing based on the knowledge accrued so far to add or remove treatments during an ongoing optimisation process. Hence finding the optimal spot along that trade-off would enable human operators to maximise yield, while also generating meaningful interpretations -- backed by sound statistical evidence, garnering more efficiency in the testing process.

\subsection{Non-stationary dynamics}
There's an important and common challenge we have to consider when applying this research in practice, to do with the fact that, in the real world, circumstances can change. Hence we need our learning system to be able to adapt to changes in underlying value of the alternatives available. Furthermore, as mentioned earlier we want this adaptation to changing circumstances to be interpretable, as we want human operators to trust, understand, and adjust the system’s behaviour.

The reason non-stationarity poses a challenge is that in the face of changing circumstances our agent has to decide what to remember and what to forget. If it remembers everything its memory would become crystallised and unable to adapt. On the flip side, if our agent only has a relatively short-term memory it will adapt quickly but will not be stable. In the worst case it might not be able to learn about real differences between the alternatives. Hence, we need an agent that adapts how much it remembers and forgets depending on the non-stationary dynamics it observes.

\section{Related Work}
Previous research has developed extensions of multi-armed bandit agents to deal with various forms of non-stationary dynamics. For example, the classical UCB and Thompson sampling agents have been extended to non-stationary settings, by either discounting past rewards, or using a fixed length sliding window memory over recent rewards \cite{Garivier2008OnProblems, Raj2017TamingApproach, Trovo2020Sliding-WindowSettings}. While these approaches can adapt to non-stationary rewards, their empirical application requires the specification of a somewhat arbitrary hyperparameter like the discounting factor or the length of the sliding window. Setting these hyperparameters correctly requires us to make a priori assumptions of the temporal dynamics of the non-stationary rewards we are likely to observe, which can be challenging to know in practice.

An alternative set of approaches have employed techniques from research into change-point detection and concept drift to dynamically adapt the length of the sliding window memory based on analysis of the statistical properties of past rewards \cite{Cavenaghi2021, Fouche2019ScalingAlgorithms, Mellor2013ThompsonDetection}. Our proposal here aligns with these approaches, in that we propose an alternative method of dynamic memory adaptation – one that targets statistical power for delivering interpretable adaptation.

\section{Contribution}
Previous research discussed above has developed algorithmic solutions to managing the agent's memory with a sole focus on minimising regret. We take a different approach: we start with the objective of providing interpretable results so as to facilitate higher-order decision-making, alongside minimising regret. What do we mean by ``interpretable''? Statistical hypothesis testing is a gold standard for evaluating evidence, and we use this notion here. The agent’s behaviour is interpretable if it can generate evidence of statistically correct behaviour alongside optimising for the maximum reward -- and if that interpretation does not suffer large changes between updates. To achieve this, we conduct hypothesis testing with Bayes Factors and adapt memory to maintain a desired set point of statistical power. We now explain our approach in formal detail.

We focus on the batched learning scenario common in real-world deployments of bandits, where the agent pulls a batch of arms between consecutive learning updates. Further, we restrict ourselves to the Thompson sampling agent \cite{Agrawal2012AnalysisProblem}, which is well-suited to the batched learning scenario because of its stochastic allocation strategy. Between consecutive updates, our agent allocates treatment units $i$ to treatment arms $a_i \in {1..n}$, selecting from one of $n$ treatment arms, using Thompson sampling over their posterior reward distributions.

\subsection{Statistical Significance using Bayesian Hypothesis Testing}
Given a pair of arms, $a_i$ and $a_j$, we evaluate the relative evidence for the hypotheses $H0$: that the true rewards of $a_i$ and $a_j$ are statistically indistinguishable, and $H1$:  that the true rewards differ by at least a certain minimum detectable effect size.

At each update, we calculate this evidence in the form of sequential Bayes factors between every pair of arms \cite{Schonbrodt2017SequentialDifferences}, which are interpreted as per \cite{Kass1995BayesFactors}:

\begin{equation}
    BF_{a_i, a_j} = \frac{p(\lvert r_{a_i}^{JS} - r_{a_j}^{JS} \rvert \;\vert H1)}
              {p(\lvert r_{a_i}^{JS} - r_{a_j}^{JS} \rvert \;\vert H0)},
\end{equation}

\noindent where $r^{JS}$ are the empirical Bayesian estimates of arm rewards up to that update, estimated using James-Stein shrinkage \cite{Efron1975DataGeneralizations}. Empirical Bayesian estimation is a robust, data-driven approach to reduce the risk of multiple comparisons \cite{Dimmery2019ShrinkageExperiments, Chennu2021SmoothFeedback}.

\subsubsection{Dynamic Setting of the Alternative Hypothesis}
The estimation of Bayes factors as described above requires the specification of the null ($H_0$) and the alternative ($H_1$) hypotheses. $H0$ and $H1$ are initialised with prior expectations about the difference between the true rewards of $a_i$ and $a_j$ if available. If not, we set them with Gaussian noise priors \cite{Dienes2014}.

When the true rewards are distributed binomially – a common use-case in industry – we exploit tools from the frequentist space to select the mean of our likelihood priors. Namely, we choose it to be the minimal difference that a correctly-powered fixed horizon test would detect with the same data. This choice derives from the idea that a fixed-horizon hypothesis test is, somehow, optimal for a given volume of data.

\subsection{Dynamic Memory Based on Statistical Power}
It has been previously shown that a Bayes factor of $K$ between a pair of arm rewards corresponds to a frequentist false discovery rate $p_d = \frac{1}{K + 1}$ \cite{Deng2016ContinuousTesting}. Inverting this relationship, we can convert an operator-specified false discovery rate of, say 0.05, to a Bayes factor threshold of 19 ($K = \frac{1}{p_d} - 1 \therefore 19 = \frac{1}{0.05} - 1$).

Importantly, this threshold can not only be used to reject $H0$ when the Bayes factor is greater than 19, but also to \emph{accept} $H0$ when the Bayes factor is less than $\frac{1}{19}$. Once we achieve this level of statistical power, we can assert that we have sufficient evidence to declare that either that the arms are significantly different ($H1$ is more likely), or equivalent (when $H0$ is more likely), at the desired level of statistical confidence specified by $p_d$. We use this insight to adapt our memory length to directly target this desired set point of statistical confidence. In other words, we keep only enough data to confidently assert a difference or equivalence between arms -- no more. 

The intuition behind our approach is as below:

\begin{enumerate}
    \item Whenever we update our agent with a new batch of data, we conduct Bayesian hypothesis tests to compare rewards of arms.
    \item If we don’t have sufficient statistical power to decide whether the arm rewards are different or equivalent, we grow our memory over updates. Over updates, having this longer memory of the past data increases statistical power.
    \item Eventually, once we have sufficient statistical power to decide either way, we start forgetting historical rewards. Naturally, this reduction in memory now decreases statistical power.
    \item This feedback loop between statistical power and memory length eventually stabilises into a dynamic equilibrium where the memory length is just sufficient to detect real differences at a desired level of statistical power, but is no more than necessary.
\end{enumerate}

More formally, at each update, given a desired false discovery rate $p_d$ (typically 5\%), Bayes factors $BF_{a_i, a_j}$ between every pair of arms, and a current memory length of $m$ recent updates, we adjust $m$ as shown in Algorithm \ref{alg:BayesWin}.

\begin{algorithm}[hbt!]
\caption{BayesWin}
\label{alg:BayesWin}
\begin{algorithmic}
\Require $0 < p_d < 1$
\State Set $K := \frac{1}{p_d} - 1$
\If{$\exists (i, j) \mid BF_{a_i, a_j} > K $}
    \State $m \gets m - 1$
\ElsIf{$BF_{a_i, a_j} < \frac{1}{K} \: \forall \: (i, j)$}
    \State $m \gets m - 1$
\Else
    \State $m \gets m + 1$
\EndIf 
\end{algorithmic}
\end{algorithm}

\section{Simulations}

\subsection{Thompson Sampling with Batched ADWIN}

To empirically evaluate the performance of BayesWin, we compare it to ADWIN \cite{Bifet2007LearningWindowing}, a popular change-point detection algorithm that has recently been shown to be an effective handle for non-stationarity in conjunction with Thompson Sampling \cite{Fouche2019ScalingAlgorithms}. In its canonical formulation, ADWIN accepts a stream of values ${x_{1}, x_2, ...}$ that it stores into memory. It continually expands this memory until the expected reward of any two contiguous subwindows of memory is significantly different, according to a statistical test with confidence $\delta$.

To enable a comparison to the batched BayesWin algorithm, we adapt the standard ADWIN algorithm to operate in the batched setting. This batched version of ADWIN behaves similarly to its standard formulation. At each update, given a change-point detection false positive rate $\delta$ and a window $W$ of recent updates with current length $m$, we adjust $m$ as shown in Algorithm \ref{alg:ADWIN}. 

\begin{algorithm}[hbt!]
\caption{BatchedADWIN}
\label{alg:ADWIN}
\begin{algorithmic}
\Require $0 < \delta < 1$
\State $m \gets m + 1$ (i.e., add the most recent batch to the window)
\Repeat
    \State $m \gets m - 1$
    \Until $| r_{W_0} - r_{W_1} | < \epsilon \: \forall \: W = W_0 \cdot W_1$
\end{algorithmic}
\end{algorithm}

We apply three tweaks to the algorithm to ensure comparability to BayesWin. First, we define it to maintain a explicit window of aggregated responses and assignments per batch, rather than of non-aggregated values. Second, the maximum likelihood estimate of the reward, $r_{W_i}$, is taken to be the weighted average of observed batched rewards, rather than a naive mean of observed non-aggregated values. In the case of binomially distributed rewards, the focus of our simulations, it is:

\begin{equation}
    r_{W_i} = \frac{\text{Total positive responses in } W_i}{\text{Total assignments in } W_i}.
\end{equation}

\noindent Third, we alter the definition of $\epsilon$, the threshold at which a change is detected, to handle the nature of the batched updates. We take the same normal distribution approximation for $\epsilon$ as in \cite{Bifet2007LearningWindowing}:

\begin{equation}
    \epsilon = \sqrt{\frac{2}{\Tilde{m}} \cdot \sigma_{W}^{2} \cdot \ln{\frac{2}{\delta'}}} + \frac{2}{3 \Tilde{m}} \ln{\frac{2}{\delta'}},
\end{equation}
where $\Tilde{m}$ is the harmonic mean of the lengths of the two subwindows of memory. But we emphasise that $\delta' = \delta / \ln |m|$, i.e., the number of \emph{batches} within the memory. This setting of $\delta'$ contrasts with standard ADWIN, which corrects for false positives by considering the total number \emph{assignments} within the memory. Additionally, in the case of binomially distributed inputs, we set $\sigma_{W}^{2} = r_{W} (1 - r_{W})$, where $r_W$ is as defined above.

\subsection{Simulation Framework}

We compare the performance of BayesWin to BatchedADWIN across two common, non-stationarity benchmarks using simulations: abrupt change, where true rewards switch at a single point in time; and gradual change, where true rewards drift gradually over updates. In both scenarios, each simulation run was repeated 100 times. At each update in each repetition, a batch of 1000 treatment units were assigned to variants, with the proportion of allocations derived from Thompson sampling over the reward distributions estimated by the BayesWin and BatchedADWIN agents.

\subsubsection{Abrupt change}

We initialise five variants to have true rewards drawn from the Beta distribution $\beta(3, 80)$ and for the first 100 updates of each simulation run these true reward values remain unchanged. Then, at the 100th update, we randomly shuffle the variants, so as to abruptly change their true rewards. For the subsequent 100 updates, we maintain these shuffled reward values.

\subsubsection{Gradual change}

We proceed with the same initial set up as in the abrupt change scenario: we initialise five variants drawn from  Beta distribution $\beta(3, 80)$. Over 300 updates, we then linearly drift the true reward of each variant towards a new value obtained by randomly shuffling the true rewards sampled at the beginning.

\subsection{Simulation Results}

\begin{figure*}[!htb]
  \centering
  \includegraphics[width=\textwidth]{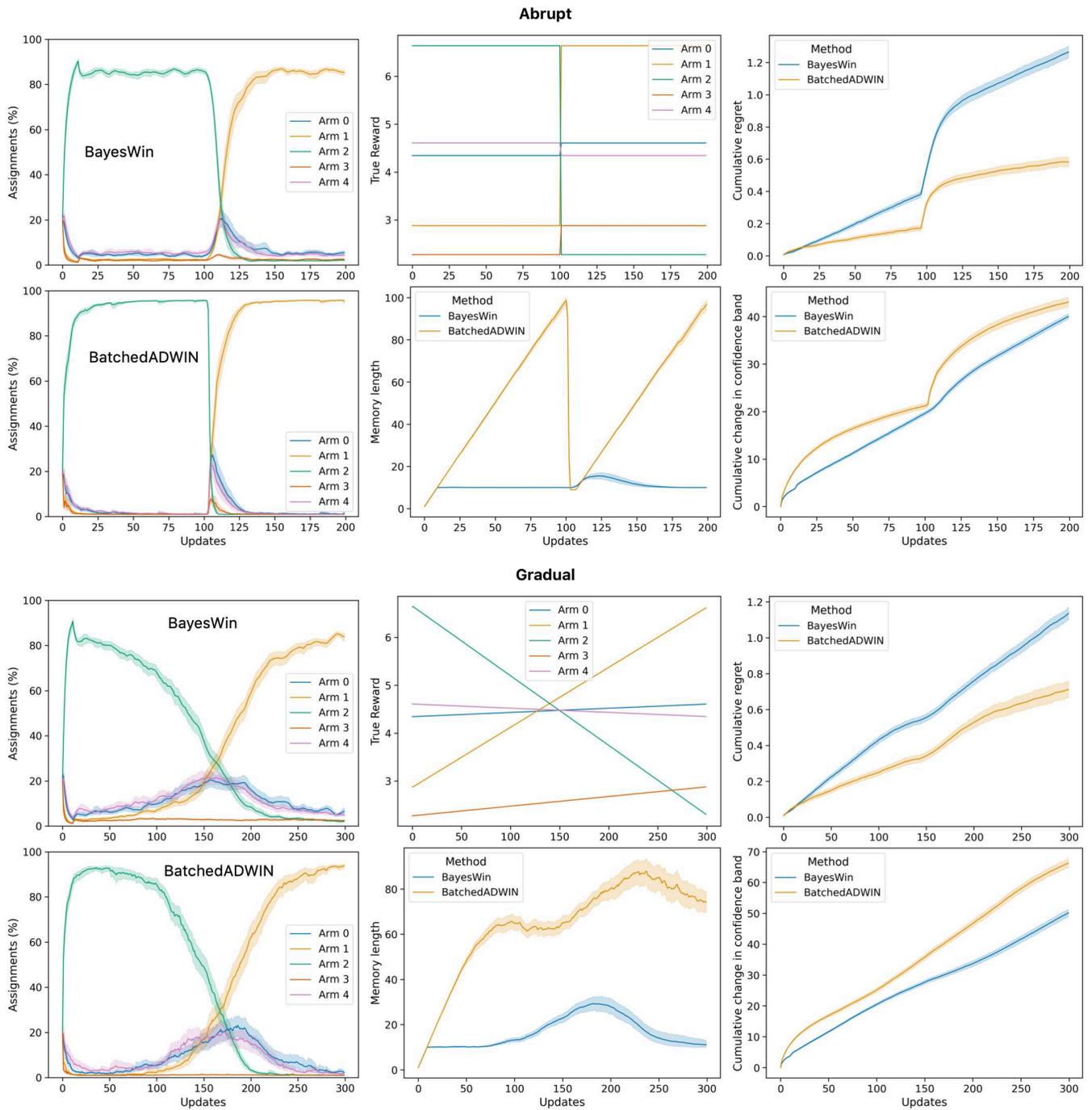}
  \caption{Simulation of BayesWin and BatchedADWIN under abrupt and gradual non-stationary conditions. Shaded area in each panel represents the 95\% bootstrap confidence interval computed over simulation runs.}
  \label{Simulations}
\end{figure*}

Fig. \ref{Simulations} compares BayesWin and BatchedADWIN under two forms of non-stationarity -- abrupt and gradual. In these simulations, the detection threshold of both algorithms -- that is, the false discovery rate $p_d$ of BayesWin and the false positive rate $\delta$ of BatchedADWIN -- were set to 0.05. We compare the:
\begin{itemize}
    \item progression of treatment assignments,
    \item length of the agent's memory,
    \item cumulative regret, and
    \item cumulative movement of sequential Bayes factors between the statistical confidence bands proposed by Kass and Raftery \cite{Kass1995BayesFactors}.
\end{itemize}

As can be seen, both methods show evidence of adaptation to the changing circumstances of the simulated scenario. However, there is a trade-off between stable interpretability and regret minimisation. In both scenarios, BayesWin produces smaller movements between confidence bands. In this sense, it delivers a more stable interpretation. Even at the point of non-stationarity, it adapts more steadily -- that is, it does not suddenly change its mind. This latter point is even more prominent in the Abrupt scenario: although BayesWin is able to identify the new best variant its change in confidence never shows an obvious discontinuity, in direct contrast to BatchedADWIN. We posit that a human operator would find smoother changes in confidence more interpretable. In contrast, from a pure regret minimisation perspective, BayesWin performs worse. Particularly in the case of abrupt change, it requires a larger number of updates before it shifts assignments to the new best performing variant.

\begin{figure}[hbt!]
  \centering
  \includegraphics[width=\linewidth]{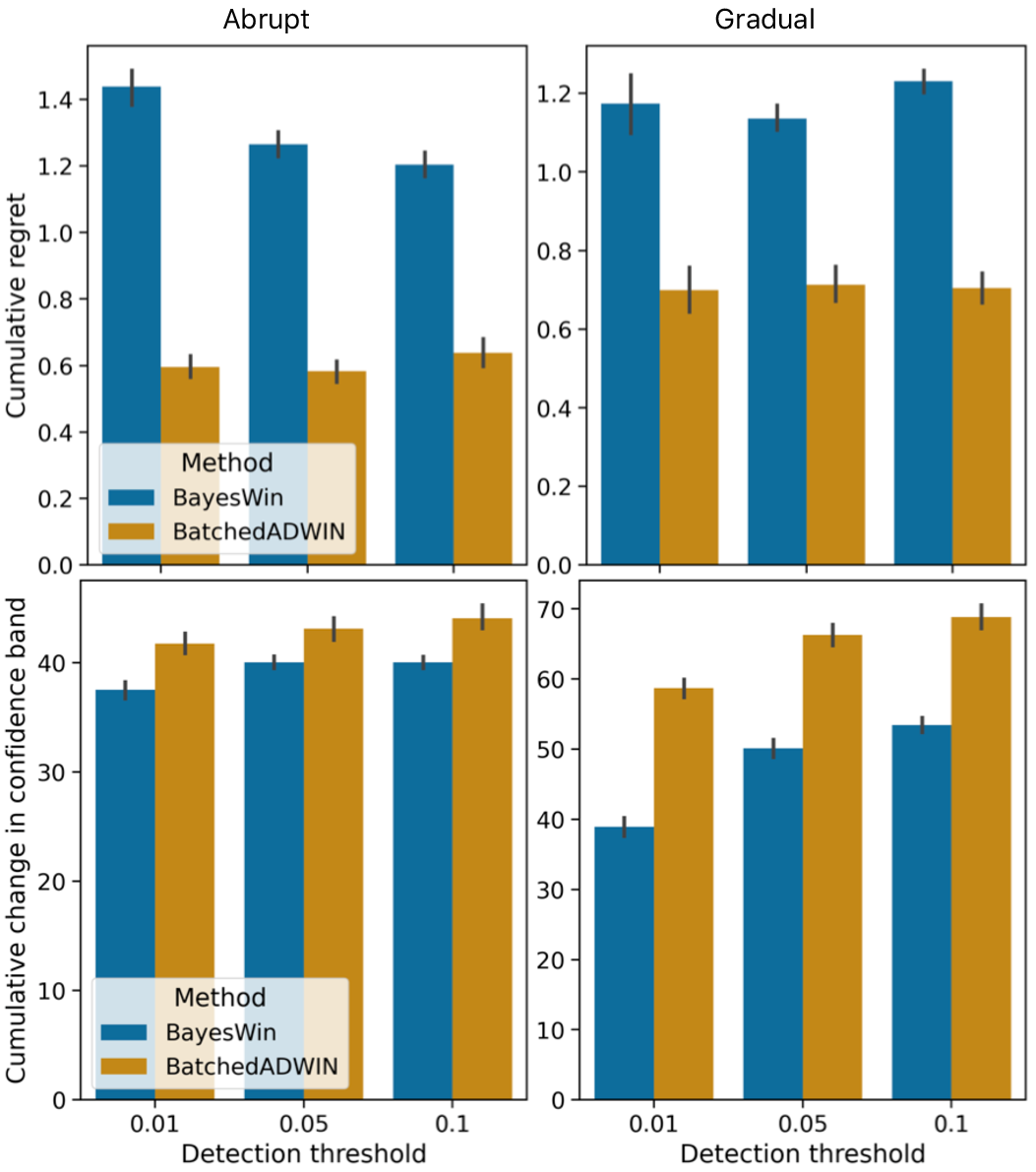}
  \caption{Comparison of BayesWin and BatchedADWIN under varying detection thresholds ($p_d$ and $\delta$ respectively). Error bar represents the 95\% bootstrap confidence interval computed over simulation runs.}
  \label{Hyperparam}
\end{figure}

To confirm the robustness of our results when varying the detection threshold, we verified that the maximal differences observed in cumulative regret and change in confidence generalised to other values of the false discovery rate ($p_d$) of BayesWin and false positive rate ($\delta$) of BatchedADWIN (see Fig. \ref{Hyperparam}). Across these simulations, BatchedADWIN was 40-60\% lower in terms of cumulative regret, and 10-50\% higher in terms of cumulative change in confidence bands. In particular, the difference in cumulative change in confidence bands was more prominent in the gradual change scenario.

\section{Real-world Application}

\subsection{Architecture}

\begin{figure}[hbt!]
  \centering
  \includegraphics[width=\linewidth]{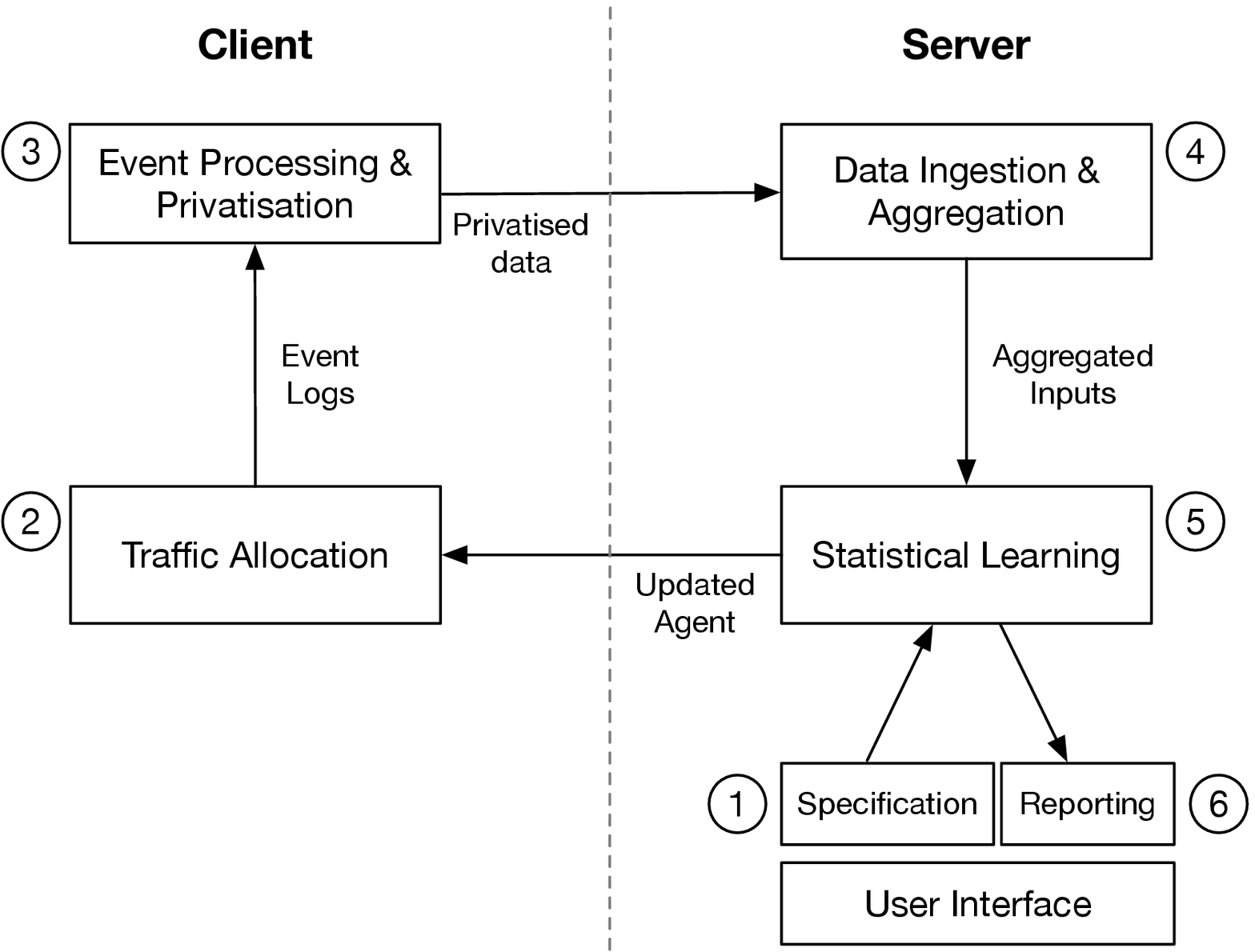}
  \caption{System architecture}
  \label{fig: Architecture Diagram}
\end{figure}

We tested the BayesWin algorithm at scale to enable business operators to optimise aspects of on-device user experience using our optimisation-as-a-service. Due to the focus on interpretability, our proposed system architecture is designed to permit the simultaneous testing of both A/B tests and multi-armed bandits as two options along a spectrum of learning designs. Fig. \ref{fig: Architecture Diagram} outlines the high-level organisation of this system -- which comprises five components that form a feedback loop -- including the points at which business operators interface with it.

\begin{figure*}[hbt!]
  \centering
  \includegraphics[width=\textwidth]{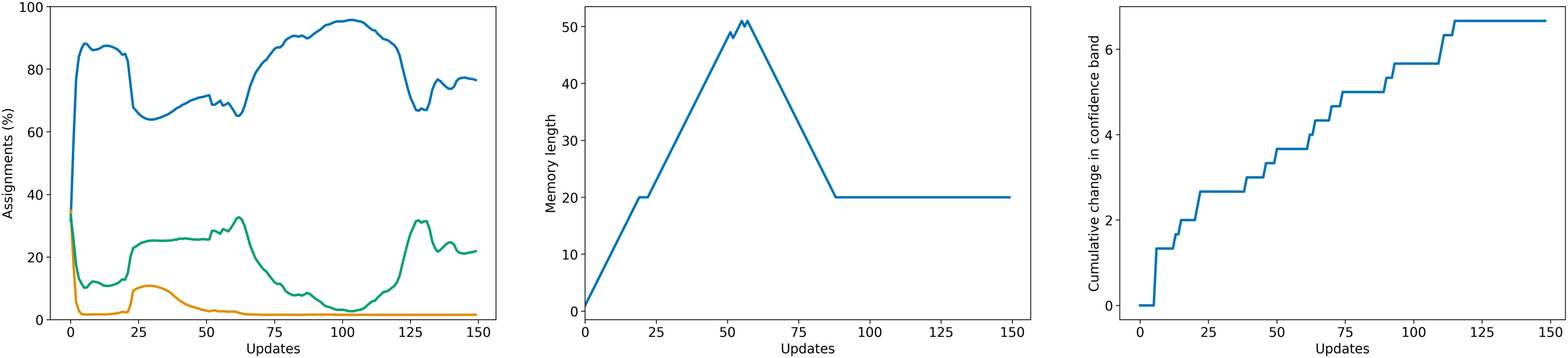}
  \caption{Real-world results from an agent incorporating BayesWin memory}
  \label{fig: Deployment Results}
\end{figure*}

\subsubsection{Learning configuration}

The first step in running a fixed allocation A/B test or a dynamic allocation multi-armed bandit requires the following specification (Fig. \ref{fig: Architecture Diagram}, Box 1):
\begin{itemize}
    \item How many variants are in the experiment
    \item Which “policy” is being run (for example, is this an A/B test or a bandit, are there stopping rules, etc.)
    \item An identifier for the success criteria or metric that the bandit will aim to optimise.
\end{itemize}

\noindent This specification is registered via an online live API service.

\subsubsection{Traffic allocation and randomisation}

The traffic allocation component is an online system running on the client device (Fig. \ref{fig: Architecture Diagram}, Box 2). It accepts a treatment unit as input and a variant allocation as output.  This traffic allocation uses a deterministic hashing algorithm that provides optimal uniqueness, speed, and uniformity properties. For each experiment the treatment unit identifier is concatenated with an experiment-level identifier to provide deterministic outputs. The output is divided uniformly and then mapped to a variant by the adaptive Bayesian agent described above. The agent initially allocates clients randomly to variants, but shifts traffic allocation over subsequent updates (see below).

\subsubsection{Local event processing and privatisation} 

In order to ensure the protection of client data, we aim to handle traffic allocation, response processing and logging locally on the client device, and anonymise any data before sending it to a central server or system (Fig. \ref{fig: Architecture Diagram}, Box 3). Hence information about the treatment variant to which the client is assigned, and the attribution of success signals, is private to the client device. Following this on-device measurement of signals on the client, we apply a range of privatisation methods:

\begin{itemize}
    \item Redaction of fields, such as identifiers
    \item De-resolution of information, such as bucketing of timestamps
    \item Injection of noise, through local differential privacy
\end{itemize}

Finally, the anonymised data are sent to a central service to perform ingestion and aggregation of the signals.

\subsubsection{Data ingestion and aggregation pipeline} 

Events are ingested as streams and delivered to a dedicated Kafka topic. Events are extracted from Kafka to a persistent data storage (for example, cloud object storage) for further transformation and aggregation by scheduled data pipelines executing Apache Spark jobs (Fig. \ref{fig: Architecture Diagram}, Box 4). The output data of the batch job consists of a count of new assignments to each variant since the last update of the agent, and new success signals as defined by the objective function.

\subsubsection{Statistical learning component}

The statistical learning component is offered both as an online service and library (Fig. \ref{fig: Architecture Diagram}, Box 5). The trade-off between these options depends on the input data volume and processing time. The adaptive Bayesian agent algorithm described above runs within this component, and provides robust reward estimates, credible intervals around these estimates, dynamic traffic allocation guidance and pairwise hypothesis testing outputs. After each update with a new batch of data, the updated agent is propagated back to the on-device traffic allocation component above. This consequently changes future treatment allocations of clients to variants, and thereby closes the feedback loop. Results of statistical inference, in the form of sequential Bayes factors and confidence bands, are reported to business operators involved in higher-order decision making based on the reported outputs of the statistical inference (Fig. \ref{fig: Architecture Diagram}, Box 6).

\subsection{Large-scale Testing Results}

Fig. \ref{fig: Deployment Results} shows results of a test of an agent incorporating the BayesWin algorithm using the architecture described above. The test included three variants that represented different alternatives of an aspect of user experience. The business operator aimed to maximise a conversion-like metric, and apply the insights from this test in the design of future treatment alternatives.

As shown in the figure, over 150 daily updates of the agent, the test aggregated privatised feedback from clients over the course of approximately five months. Similar to the simulation results in Fig. \ref{Simulations}, we show the evolution of treatment allocations, the length of the agent's dynamic memory at each update, and the cumulative change in the statistical confidence bands.

From early in the test, it is clear that the bandit identifies one variant as being dominant. Nonetheless there is a degree of non-stationarity in the assignment of treatments to different variants: the share of assignments to the top two variants ebbs and flows over the long period of test, potentially owing to cyclical dynamics in the true rewards. The BayesWin agent responds to this non-stationarity -- it gradually expands its memory for the first twenty updates, to reach a minimum length. At that point, the statistical confidence associated with each pair of branches is lower than the threshold: hence it continues to expand memory until the fifty-fifth update. At that point, the data is sufficient to suggest that one variant is statistically the best. The memory shrinks, in response, until it returns to its minimum length, and maintains at that level, enabling the agent to remain ready to adapt to cyclical changes in rewards. Alongside, the business operator is shown statistical confidence in significant differences between the variants, for the purposes of interpretation and decision making. Importantly, as can be seen in Fig. \ref{fig: Deployment Results}, the cumulative movement of this statistical confidence between confidence bands remains well bounded, aiding interpretability.

\section{Discussion and Further Work}

In real-world settings and applications, we desire systems which can assist human operators with complex optimisation problems. It is therefore important to design algorithms that can work in collaboration with human operators, with the algorithm providing useful feedback, even if this incurs slightly higher regret. To this end, we have developed BayesWin, which aims to balance the trade-off between regret and interpretability. While existing proposals in the literature are successful at the task of minimising regret, they can produce behaviour that is harder to interpret. When viewing their results through the lens of statistical hypothesis testing, they can generate statistical confidence values that suddenly drop dramatically from one update to the next.

This trade-off between regret minimisation and interpretability is borne out by our simulations results. The statistical confidence of BayesWin does, indeed, evolve more steadily than that of ADWIN. But there is no free lunch - this boost to interpretability comes at the cost of higher regret. The optimal point along this trade-off depends on the deployment context, and there are indeed real-world situations where interpretable solutions that sacrifice some regret might be preferable. The BayesWin algorithm articulated here represents one such solution along this trade-off, and future research should explore additional solutions in this space.

\balance

\bibliographystyle{ACM-Reference-Format}
\bibliography{references}
\end{document}